\documentclass{article}
\usepackage{ijcai22}

\usepackage{times}
\usepackage{soul}
\usepackage{url}
\usepackage[hidelinks]{hyperref}
\usepackage[utf8]{inputenc}
\usepackage[small]{caption}
\usepackage{graphicx}
\usepackage{amsmath}
\usepackage{amsthm}
\usepackage{booktabs}
\usepackage{algorithm}
\usepackage{algorithmic}
\usepackage{amssymb}
\usepackage{subcaption}
\def\clip{\operatorname{clip}}
\usepackage{natbib}
\usepackage{multirow}
\DeclareMathOperator*{\argmax}{arg\,max}

\newtheorem{theorem}{Theorem}

\title{Look Before Leap: Look-Ahead Planning with Uncertainty in Reinforcement Learning}

\author{
Yongshuai Liu
\And Xin Liu
\affiliations
University of California, Davis
\emails
\{yshliu, xinliu\}@ucdavis.edu,
}

\begin{document}

\maketitle

\begin{abstract}
  Model-based reinforcement learning (MBRL) has demonstrated superior sample efficiency compared to model-free reinforcement learning (MFRL). However, the presence of inaccurate models can introduce biases during policy learning, resulting in misleading trajectories. The challenge lies in obtaining accurate models due to limited diverse training data, particularly in regions with limited visits (uncertain regions). Existing approaches passively quantify uncertainty after sample generation, failing to actively collect uncertain samples that could enhance state coverage and improve model accuracy. Moreover, MBRL often faces difficulties in making accurate multi-step predictions, thereby impacting overall performance. To address these limitations, we propose a novel framework for uncertainty-aware policy optimization with model-based exploratory planning. In the model-based planning phase, we introduce an uncertainty-aware k-step lookahead planning approach to guide action selection at each step. This process involves a trade-off analysis between model uncertainty and value function approximation error, effectively enhancing policy performance. In the policy optimization phase, we leverage an uncertainty-driven exploratory policy to actively collect diverse training samples, resulting in improved model accuracy and overall performance of the RL agent. Our approach offers flexibility and applicability to tasks with varying state/action spaces and reward structures. We validate its effectiveness through experiments on challenging robotic manipulation tasks and Atari games, surpassing state-of-the-art methods with fewer interactions, thereby leading to significant performance improvements.
\end{abstract}

\section{Introduction}\label{sec:intro}
Model-based Reinforcement Learning (MBRL) methods have emerged as powerful approaches that exhibit sample efficiency compared to model-free methods~\citep{liu2021policy}, both in theory and practice~\citep{liu2024towards,janner2019trust}. MBRL leverages an approximate dynamics model of the environment to aid in policy learning~\citep{liu2023constrained}. However, acquiring a precise model, which plays a critical role in MBRL, presents a challenge, particularly in real-world applications~\citep{liu2021cts2,liu2021resource,liu2020constrained}. An inaccurate model can introduce biases during policy learning, ultimately leading to the generation of misleading trajectories. This prevalent issue is commonly referred to as the model bias problem~\citep{deisenroth2011pilco}.

Attaining an accurate model presents challenges due to the continuous updates of the policy, causing a shift in the distribution over visited states. The modified distribution often lacks of enough diverse training data for effective approximation, which can result in inaccurate model, particularly in regions with limited visits (uncertain regions). To mitigate this challenge, existing methods employ uncertainty-aware dynamic model techniques~\citep{halev2024microgrid}, such as utilizing a bootstrapped model ensemble~\citep{chua2018deep} or Bayesian RL~\citep{liu2022farsighter}. However, these approaches passively quantify uncertainty only after the samples have been generated, failing to actively collect uncertain samples that could offer a broader coverage of states and enhance the model's accuracy.

Moreover, an additional challenge in MBRL is that the learned models often involve predicting multiple steps ahead, such as Model Predictive Control~\citep{camacho2013model}. However, the learning objective of the model primarily focuses on minimizing one-step prediction errors. Consequently, an objective mismatch~\citep{lambert2020objective} arises between model learning and model usage. While policy optimization relies on trajectories with multi-step predictions (model usage), the model is primarily trained to prioritize the accuracy of immediate predictions (model learning). Although it is possible to generate multi-step predictions using one-step models by iteratively feeding the predictions back into the learned model, this approach can lead to deviations from the true dynamics. Accumulated errors, stemming from the model's lack of optimization for multi-step predictions, can impact the accuracy of multi-step predictions and overall performance.

To overcome the limitations mentioned above, our paper presents a novel framework for uncertainty-aware policy optimization with model-based exploratory planning. In the  model-based planning phase, we introduce an uncertainty-aware k-step lookahead planning mechanism to guide action selection at each step. This planning process operates under an approximate uncertainty-aware model and an approximate value function. 
We not only present a bound for multiple-step prediction that explicitly factors in both the model uncertainty and the value function approximation error, but our analysis also uncovers an inherent trade-off between these two aspects. Furthermore, our empirical experiments demonstrate leveraging this trade-off can enhance policy performance in our framework.
Specifically, during action selection at each step, we simulate k-step roll-outs and sample a number of `fantasy' samples utilizing the uncertainty-aware model. This simulated sampling procedure enables us to construct a tree structure, facilitating the selection of action sequences that result in the highest accumulated reward within a k-step rollout. We then select the first-planned action as the action to be executed in real environment in each step.

In the policy optimization phase, we leverage an uncertainty-driven exploratory policy~\citep{burda2018exploration} to actively collect diverse training samples from the environment. This approach mitigates the detrimental effects of insufficient exploration on both policies and forward dynamics models. By gathering more informative data through this exploratory policy, we enhance the learning accuracy of our models, which in turn improves the overall performance of our RL agent.

Overall, we make several significant contributions:
\begin{itemize}
    \item 
     We introduce uncertainty-aware model-based k-step lookahead planning. The proposed Theorem bound for multi-step planning explicitly accounts for both model uncertainty and value function approximation error, revealing an inherent trade-off between these factors. Unlike the prevalent literature, which focuses on one-step uncertainty in MBRL, our approach provides a broader perspective.
    \item We demonstrate the effectiveness of using uncertainty-driven exploration in  policy optimization to collect training data, improving the quality of the dynamics model beyond the conventional goal of enhancing policy performance. Our approach mitigates the issue of insufficient exploration on policies and forward dynamics models, representing a novel usage.
    \item We propose a novel framework that actively incorporates uncertainty exploration in both model-based planning  and policy optimization phases. This approach diverges from existing literature, which typically addresses uncertainty on only one side or integrates both aspects without considering uncertainty.
    \item 
    Our approach does not rely on assumptions of perfect forward dynamics models or state/action spaces, making it more scalable and applicable to a broader range of tasks. We evaluate our method on challenging tasks in MuJoCo and Atari games. The experimental results demonstrate that our approach outperforms SOTA methods with less data. Furthermore, the results show that our method is scalable across a wide range of RL tasks, handling high/low-dimensional states, discrete/continuous actions, and dense/sparse rewards.
\end{itemize}

\section{Related Work}\label{sec:related}

\subsection{Uncertainty-aware MBRL}
One of the key challenges in MBRL is effectively handling uncertainty associated with our model predictions. By doing so, we can identify instances where our predictions may be less reliable when planning based on our model. In statistics, there are two main approaches to uncertainty estimation: frequentist and Bayesian methods.
The frequentist approach, as demonstrated by~\citep{chua2018deep}, employs techniques such as statistical bootstrapping for model estimation. For example, Plan2Explore~\citep{sekar2020planning} estimates state uncertainty with latent disagreement and generates model training data with exploration policy. Bayesian RL methods have been extensively surveyed by~\citep{ghavamzadeh2015bayesian}. Non-parametric Bayesian methods, including Gaussian Processes, have shown great success in modeling estimation, as exemplified by PILCO~\citep{deisenroth2011pilco}. However, GPs suffer from computational scalability issues when dealing with high-dimensional state spaces.
Techniques like variational dropout~\citep{mcallister2016improving} and variational inference~\citep{depeweg2016learning} have been applied to neural network-based Bayesian modeling of dynamics. These approaches aim to address the limitations of GPs and provide efficient uncertainty estimation in high-dimensional state spaces.
A recent study, COPlanner~\citep{wang2023coplanner}, shares similarities to our method. It utilizes an Bayesian-based uncertainty-aware policy-guided model predictive control (UP-MPC) component for multi-step uncertainty estimation in planning. This estimated uncertainty is then applied as a penalty during model rollouts and as a bonus during exploration in the real environment to guide action selection. We will discuss the advantages of our approach over these methods in Sec.~\ref{sec:k-step} and Sec.~\ref{sec:exp}.

\subsection{Uncertainty-driven Exploration in MFRL}
Uncertainty-driven exploration considers environmental uncertainty and can be classified into two categories.
The first category is intrinsic motivation methods, which augment the \textbf{extrinsic reward} from the environment with an exploration bonus, known as \textbf{intrinsic reward}. This additional reward incentivizes the agent to explore states that are more uncertain. 
One approach for estimating intrinsic rewards is count-based exploration, such as~\citep{ostrovski2017count}. These methods estimate the number of times a state has been visited, denoted as $n(s)$, and use it to compute the intrinsic reward. Common formulations include $1/n(s)$ or $1/\sqrt{n(s)}$.
Another approach is prediction-based exploration, such as the RND algorithm introduced by~\citep{burda2018exploration}. Prediction-based methods estimate state novelty based on prediction error. The underlying assumption is that if similar states have been encountered frequently in the past, the prediction error for a given state should be lower~\citep{liu2023adventurer}.
The second category is uncertainty-oriented methods, as discussed in works by~\citep{azizzadenesheli2018efficient}. These methods typically model uncertainty using the Bayesian posterior of the value function to capture uncertainty. By doing so, the agent is encouraged to explore regions with high uncertainty.

\begin{table*}[t]
\centering
\caption{Mathematical symbol list}\label{tb:symbol}
    \begin{tabular}{ |c|c|c|} 
        \hline
        \multirow{4}{5.3em}{Global}& $\pi_{\phi}$ & parametered control policy (PPO) \\
        % \cline{2-3}
          & $V_{\sigma}$ &  value function approximation\\
          % \cline{2-3}
         &$f_{\theta}$& Variational Bayes (VB) dynamic model\\
        % \cline{2-3}
        &$f_{\theta_r}$&reward function approximation\\
        \hline
         \multirow{7}{5.3em}{Policy optimization phase}&$s_t$& real environment state at time $t$\\
        % \cline{2-3}
        & $a_t$&action executed in the real environment obtained from model-based planning\\
        % & &\\
        % \cline{2-3}
        &$r_t^i$&intrinsic reward from RND at time t\\
        % \cline{2-3}
        &$A_t^i$&intrinsic reward advantage function at time t\\
        % \cline{2-3}
        &$r_t^e$&extrinsic reward from real environment at time t\\
        % \cline{2-3}
        &$A_t^e$&extrinsic reward advantage function at time t\\
        % \cline{2-3}
        &$\beta$&intrinsic-driven exploration weight\\
        \hline
        \multirow{11}{5.3em}{Model-based planning phase}&$M$&number of VB weight candidates \\
        % \cline{2-3}
        &$N$&number of simulated trajectories for each VB weight candidate\\
        % \cline{2-3}
        &$m$&the index to the m-th VB weight $\theta^m$\\
        &$n$&given a VB weight $\theta^m$, the index to the n-th trajectory\\
        &$k$&number of lookahaed step\\
       % \cline{2-3}
        &$\hat{s}_t^{m,n}$&simulated state with $f_{\theta}$ at time $t$ \\
        % \cline{2-3}
        &$\hat{a}_t^{m,n}$&simulated action with $\pi_{\phi}$ at time $t$\\
        % \cline{2-3}
        &$\hat{r}_t^{m,n}$&simulated reward with $f_{\theta_r}$ at time $t$\\
        % \cline{2-3}
        &$\hat{\tau}^{m,n}$&simulated trajectory associated with $\hat{s}_t^{m,n}$, $\hat{a}_t^{m,n}$\\
        % \cline{2-3}
        &$\hat{G}$&simulated accumulated rewards associated with $\hat{\tau}^{m,n}$\\
        % \cline{2-3}
        &$a_t^*$&first planned action for policy optimization phase\\
        \hline      
    \end{tabular}
\end{table*}

\subsection{Model-Based Policy Optimization (MBPO)}
A series of model-based policy optimization methods~\citep{gu2016continuous,feinberg2018model,janner2019trust} use learned models to generate synthetic data for policy training. However, these approaches often struggle with long-horizon tasks due to the accumulation of model errors. As data generation involves sampling from the initial state distribution and propagating forward using the learned model, errors compound over time, making accurate learning increasingly difficult. 
The authors tackles this issue by starting from random states previously encountered during interaction with the environment and propagating only short trajectories forward. This strategy mitigates error accumulation by replacing a few error-prone long trajectories with numerous shorter, more accurate ones.

In contrast, our proposed method diverges from MBPO by not using the model to generate training samples. Instead, we leverage the model for planning during action selection in the environment. Additionally, our approach incorporates uncertainty into model-based planning by limiting `fantasy' samples to a fixed number of steps.

\section{Preliminaries and Notations}
\subsection{Markov Decision Process}
A Markov Decision Process (MDP) is defined by a tuple $\left(\mathit{S},\mathit{A}, \mathit{R},\mathit{P}, \mathit{\mu}, \mathit{\gamma} \right )$ \citep{sutton2018reinforcement,liu2020ipo}. Here, $\mathit{S}$ represents the set of states $s_t$, where $t$ denotes the timestamp; $\mathit{A}$ stands for the set of actions $a_t$; $\mathit{R}$ denotes the reward function, with $r_t = R(s_t, a_t)$; $\mathit{P}$ signifies the transition probability function, with $\mathit{P}(s_{t+1}|s_t,a_t)$ representing the transition probability from state $s_t$ to state $s_{t+1}$ when action $a_t$ is taken; $\mathit{\mu}$ is the initial state distribution, and $\mathit{\gamma}$ represents the discount factor for future rewards. A policy $\pi$ is a mapping from states to a probability distribution over actions, with $\pi(a_t|s_t)$ indicating the probability of taking action $a_t$ in state $s_t$. We denote a policy $\pi$ as $\pi_{\phi}$ to highlight its dependence on the parameter $\phi$. In an MDP, a common objective is to determine a policy $\pi_{\phi}$ that maximizes the discounted cumulative reward. It is denoted as
\begin{equation*}
    \max_{\phi}~J^{\pi_{\phi}} = \mathbb{E}_{\tau \sim \pi_{\phi}}[\sum_{t=0}^{\infty}\mathit{\gamma}^{t}\mathit{R}(s_{t},a_{t})],
\end{equation*}
where $\tau = (s_{0}, a_{0},s_{1}, a_{1}... )$ denotes a trajectory, and $\tau \sim \pi_{\phi}$ means that the distribution over trajectories is following policy $\pi_{\phi}$. For a trajectory starting from state $s$,  the value function is
$
    V{(s)} = \mathbb{E}_{\tau \sim \pi_{\phi}}[\sum_{t=0}^{\infty}\gamma^{t}\mathit{R}(s_{t},a_{t})|s_{0}=s].
$
The  Q-function of state $s$ and action $a$ is 
$
    Q{(s,a)} = \mathbb{E}_{\tau \sim \pi_{\phi}}[\sum_{t=0}^{\infty}\gamma^{t}\mathit{R}(s_{t},a_{t})|s_{0}=s, a_{0}=a]
$
and 
the advantage function is computed as the difference between the Q-function and the value function, represented by
\begin{equation}~\label{eq:advantage}
    A{(s,a)} = Q{(s,a)} - V{(s)}.
\end{equation}

\subsection{Proximal Policy Optimization}
  Proximal Policy Optimization (PPO)~\citep{schulman2017proximal,liu2021clara} approximates the objective by a first-order surrogate optimization problem to reduce the complexity of TRPO~\citep{schulman2015trust}, defined as 
\begin{equation}\label{eq:ppo}
\resizebox{\hsize}{!}{$
\begin{aligned}
    &\max_{\phi}~L^{CLIP}(\phi ) =  \\
    &\mathbb{E}_{s_t,a_t \sim \pi_{\phi}}[\min(r_{t}(\phi)A(s_t,a_t)),
\clip(r_{t}(\phi),1-\epsilon ,1+\epsilon)A(s_t,a_t))],
\end{aligned}
$}
\end{equation}
where $r_t(\phi) = \frac{\pi_{\phi}(a_{t}|s_{t})}{\pi_{\phi{old}}(a_{t}|s_{t})}$, $A(s_t,a_t)$ is the advantage function Eq.~\eqref{eq:advantage} at time step $t$, $\clip(\cdot)$ is the clip function and $r_{t}(\phi)$ is clipped between $\left [ 1-\epsilon, 1+\epsilon  \right ]$. 

\begin{figure*}[t]
    \centering
    \includegraphics[width=0.9\textwidth]{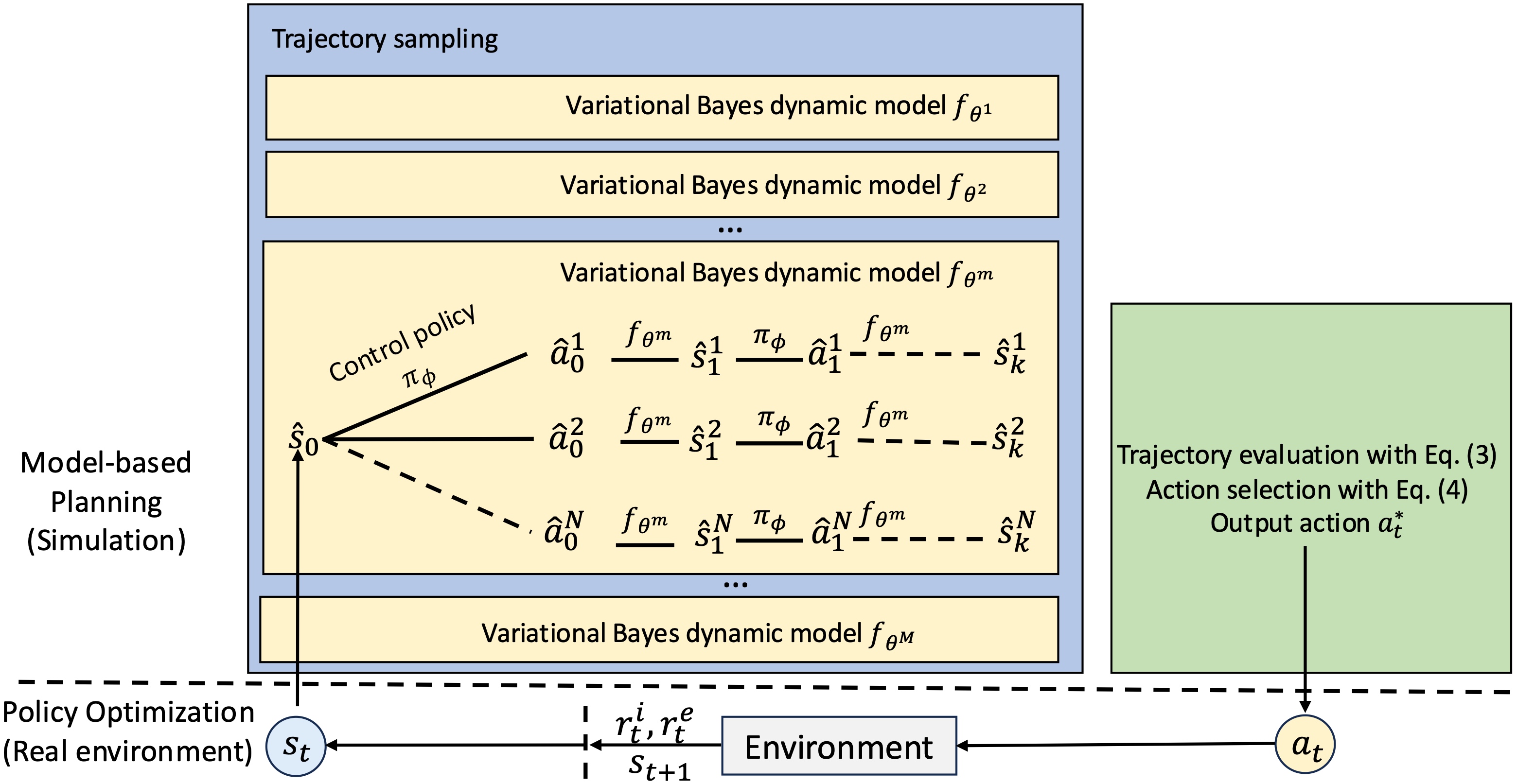}
    \caption{The illustration of the look before leap framework.}
    \label{fig:tree}
    % \vspace{-0.3cm}
\end{figure*}

\section{Lookahead Planning with Uncertainty}\label{sec:model_based}
In our work, we introduce a novel framework for uncertainty-aware policy optimization with model-based exploratory planning.
In model-based planning phase, our framework incorporates an uncertainty-aware k-step lookahead planning, which introduces a trade-off between model uncertainty and value function approximation error. 
In policy optimization phase, we leverage uncertainty-driven exploration to gather more diverse and informative samples. By actively exploring uncertain states, we enhance the learning accuracy of our dynamic models, ultimately improving the overall performance.
In Section~\ref{sec:train}, we provide a detailed explanation of our method's overall framework, including the joint learning of policy, value function, and uncertainty-aware dynamic models. Additionally, in Section~\ref{sec:k-step}, we introduce the uncertainty-aware model-based k-step lookahead planning approach. To aid in understanding the formulations, we listed the mathematical symbols in Table~\ref{tb:symbol}.

\subsection{The Framework}~\label{sec:train}
Our framework comprises two primary phases. Fig.~\ref{fig:tree} visually depicts the policy optimization and planning process, where the policy optimization phase operates in the real environment, with each action obtained from the planning phase. The model-based planning phase operates in the simulated uncertainty-aware dynamic model, selecting actions through simulated trajectory sampling, evaluation, and action selection.

Moreover, in the policy optimization phase, the agent employs an uncertainty-driven exploration as depicted in Alg.\ref{alg:framwork}. The reward at each step is formulated as $r_t = r^i_t + \beta \cdot r^e_t$.
Here, $r^i_t$ serves as an intrinsic reward designed to incentivize exploration of novel or less frequently visited states or actions. This intrinsic reward is computed using the Random Network Distillation (RND) method \citep{burda2018exploration}, which evaluates the disparity between the predictions of a randomly initialized fixed neural network (the exploration bonus network) and those of another neural network (the target network), which learns to predict the outputs of the random network. This discrepancy serves as the intrinsic reward signal.
Meanwhile, $r^e$ represents the extrinsic reward obtained directly from the real environment. The hyperparameter $\beta$ plays a role in integrating these intrinsic and extrinsic rewards, guiding the agent's learning process and striking a balance between exploration and exploitation in RL tasks.

Each action taken during the policy optimization phase is derived from the model-based planning phase (Alg~\ref{alg:framwork}: line 4), rather than relying on the policy network. In the planning phase, the agent engages in simulated k-step lookahead planning utilizing an uncertainty-aware dynamic model $f_{\theta}$ at each step. 
The agent generates virtual k-step rollout trajectories through simulation, evaluating each trajectory to facilitate the selection of action sequences that lead to the highest accumulated reward within a k-step rollout. Subsequently, we choose the first-planned action as the action to be executed for policy optimization. This action guides the agent's interaction with the real environment, with the resulting interactions appended to the training dataset (Alg:~\ref{alg:framwork}: lines 7). Further insights into the operation of the model-based planning phase will be provided in Sec.~\ref{sec:k-step}. 

Finally, the agent updates the control policy $\pi_{\phi}$, value function approximation $V_{\sigma}$, uncertainty-aware dynamics model $f_{\theta}$ and reward function approximation $f_{\theta_r}$ by sampling batches from the dataset $D$ (Alg.~\ref{alg:framwork}: lines 10-12).

\begin{algorithm}[t]
\caption{Look before leap} \label{alg:framwork}
\begin{flushleft}
number of epochs $E$, policy and model update frequency $F$, intrinsic-driven exploration weight $\beta$, PPO control policy $\pi_{\phi}$, value function $V_{\sigma}$, Variational Bayes (VB) dynamic model $f_{\theta}$, reward function approximation $f_{\theta_r}$
\end{flushleft}
\begin{algorithmic}[1]
 \FOR {epoch = 0 to E}
    \FOR {t = 0 to F}
        \STATE Current state $s_t$
        \STATE Get action from planning phase: $a_t$ = \textbf{UKP}($s_t$, $\pi_{\phi}$, $V_{\sigma}$, $f_{\theta}$, $f_{\theta_r}$)
        \STATE Execute $a_t$ in real environment, observe next real state $s_{t+1}$ and extrinsic reward $r_t^e$
        \STATE Calculate intrinsic reward $r_t^i$
        \STATE Append $\{(s_t,a_t,s_{t+1},r_t^e,r_t^i)\}$ to dateset $D$
        \STATE Calculate intrinsic reward advantages $A_t^i$ and extrinsic reward advantages $A_t^e$
    \ENDFOR
    \STATE Update PPO policy using the combined reward advantages $A_t=A_t^i+\beta*A_t^e$ with Eq.~\eqref{eq:ppo}
    \STATE Update value function approximation $V_{\sigma}$ with extrinsic rewards in $D$
    \STATE Update VB dynamic model $f_{\theta}$ (and reward function approximation $f_{\theta_r}$)  with data in $D$
\ENDFOR
\end{algorithmic}
\end{algorithm}

\begin{algorithm}[t]
\caption{Uncertainty-aware k-step lookahead planning: \textbf{UKP}($s_t$, $\pi_{\phi}$, $V_{\sigma}$, $f_{\theta}$, $f_{\theta_r}$)} \label{alg:k-step}
\begin{flushleft}
number of VB weight candidates $M$, number of simulated trajectories N for each VB weight candidate, number of lookahaed step $k$
\end{flushleft}
\begin{algorithmic}[1]
\STATE $\Phi = \left \{  \right \}$
\FOR {m = 1 to M}
    \STATE Sample VB dynamic model weight $\theta^m$
    \FOR {n = 1 to N}
        \STATE $\hat{\tau}^{m,n} = \{ \}$
        \STATE $\hat{s}_0^{m,n}$ = $s_t$
        \FOR {t = 0 to k}
            \STATE Sample action $\hat{a}_t^{m,n} \sim  \pi_{\phi}(\cdot|\hat{s}_t^{m,n})$
            \STATE Rollout VB dynamic model $\hat{s}_{t+1}^{m,n} = f_{\theta^m}(\hat{s}_t^{m,n}, \hat{a}_t^{m,n})$ and reward $\hat{r}_{t}^{m,n}$ = $f_{\theta_r^m}(\hat{s}_t^{m,n}, \hat{a}_t^{m,n})$
            \STATE $\hat{\tau}^{m,n} = \hat{\tau}^{m,n} \bigcup \{\hat{s}_t^{m,n}, \hat{a}_t^{m,n}\}$
        \ENDFOR
        \STATE $\hat{\tau}^{m,n} = \hat{\tau}^{m,n} \bigcup \{ \hat{s}_{t+1}^{m,n} \}$
        \STATE $\Phi = \Phi \bigcup \hat{\tau}^{m,n}$
    \ENDFOR
\ENDFOR
\STATE Evaluate M*N trajectories by computing accumulated rewards according to Eq.~\eqref{eq:eva}
\STATE Compute $a_{t:t+k-1}^{*}$ according to Eq.~\eqref{eq:max}
\STATE \textbf{Return}  $a_t^*$
\end{algorithmic}
\end{algorithm}

\subsection{Uncertainty-aware k-step Lookahead Planning}~\label{sec:k-step}
In model-based planning phase, we employ neural network $f_\theta$ to approximate the transition probability function $\mathit{P}(s_{t+1}|s_t,a_t)$ in a supervised manner. The network takes $(s_t, a_t)$ as input and generate $s_{t+1}$ as output. Similarly, we also utilize $f_{\theta_r}$ to approximate the reward function $\mathit{R}$. This network takes $(s_t, a_t)$ as input and produces $r_{t}$ as output.

Furthermore, to capture the uncertainty associated with the transition $\mathit{P}(s_{t+1}|s_t,a_t)$, we employ a Bayesian neural network (BNN). This BNN introduces a posterior distribution over the weights of the neural network $f_\theta$ to represent model uncertainty. However, obtaining the true posterior of a BNN is computationally challenging, particularly in complex state settings. Gal et al.~\citep{gal2016dropout} demonstrate that dropout can serve as a variational Bayesian approximation. The uncertainty in the weights is realized by applying dropout during both training and testing phases.

Using this uncertainty-aware learned model, we conduct trajectory sampling and evaluation through simulation. An action is then selected for policy optimization based on the accumulated reward and terminal state value function.

\subsubsection{Trajectory Sampling, Evaluation and action selection}
Incorporating model uncertainty is crucial for action selection to compensate for model bias. When given the current state $s_t$ from the real environment, we aim to choose an action for operation in the real environment using uncertainty-aware k-step lookahead planning.
As shown in Alg.~\ref{alg:k-step}, to incorporate the model uncertainty, we samples a set of Variational Bayes (VB) weight candidate from the VB dynamic model prior to sampling (Alg.\ref{alg:k-step}: line 3). Given a VB weight $\theta^m$, we generate $N$ actions following the distribution of the control policy: $\hat{a}_t = \{ \hat{a}_t^{m,1} , \hat{a}_t^{m,2} , ..., \hat{a}_t^{m,N}\}$. Then, for each action, we perform k-step rollout and also obtain rewards from the approximated dynamic model $f_{\theta^m}$ and reward function $f_{\theta_r^m}$ (Alg.\ref{alg:k-step}: lines 8-10). 

In the trajectory sampling stage, we simulates a set of trajectories: $$\Phi = \{ \hat{\tau}^{m,n} = [\hat{s}_0^{m,n},\hat{a}_0^{m,n},\hat{s}_1^{m,n},...,\hat{s}_{k}^{m,n}] \}_{m=1,n=1}^{M,N},$$ where $M*N$ denotes the number of simulated trajectories, $m, n$ denotes the index of a trajectory within $\Phi$, and $k$ indicates the planning horizon. $\hat{\tau}^{m,n}$ can be obtained by sequentially applying
\begin{equation*}
    \hat{s}_0^{m,n} = s_t; \hat{s}_{t+1}^{m,n} = f_{\theta^m}(\hat{s}_t^{m,n}, \hat{a}_t^{m,n}); \hat{a}_t^{m,n} \sim \pi_{\phi}(\cdot|\hat{s}_t^{m,n}),
\end{equation*} where $\hat{s}_t^{m,n}$ denotes the simulated state at planning step $t$ within $\hat{\tau}^{m,n}$, $\hat{a}_t^{m,n}$ is the action applied to $\hat{s}_t^{m,n}$, and $\pi_\phi$ denotes the control policy.  In each step, we assume that we can draw `fantasy' instances of the next state $ \hat{s}_{t+1}^{m,n}$ from the VB dynamic model $f_{\theta^m}$. The state instances are considered `fantasy' because they are simulated using the VB dynamic model $f_{\theta^m}$ rather than the real environment.

Next, in the trajectory evaluation stage (Alg.\ref{alg:k-step}: line 16), we evaluate each trajectory $\hat{\tau}^{m,n}$ with the approximated reward function $\hat{r}_{t}^{m,n} = f_{\theta_r^m}(\hat{s}_t^{m,n}, \hat{a}_t^{m,n})$ and the terminal state value function $V_{\sigma}(\hat{s}_{k}^{m,n})$, defined by:
\begin{equation}~\label{eq:eva}
    \hat{G}(\hat{s}_{0:k}^{m,n}, \hat{a}_{0:k-1}^{m,n}) = \sum_{t=0}^{k-1}\gamma^{t} \hat{r}_{t}^{m,n} +\gamma^kV_{\sigma}(\hat{s}_{k}^{m,n}).
\end{equation} 
Here, $\hat{G}(\hat{s}_{0:k}^{m,n}, \hat{a}_{0:k-1}^{m,n})$ denotes the simulated accumulated rewards associated with $\hat{\tau}^{m,n}$, and $\hat{s}_{0:k}^{m,n}, \hat{a}_{0:k-1}^{m,n}$ represent the state and action sequences extracted from $\hat{\tau}^{m,n}$, respectively. This evaluation presents a trade-off measurement between model uncertainty and value function approximation error. Our analysis in Sec.\ref{sec:why} and empirical experiments in Sec.\ref{sec:exp_lookahead} illustrate the effectiveness of leveraging this trade-off to enhance policy performance in deep RL.

Finally, in the action selection stage (Alg.\ref{alg:k-step}: lines 17-18), the first control action $a_t^{*}$ from the action sequence $a_{t:t+k-1}^{*}$ is selected based on the highest value with respect to Eq.~\eqref{eq:eva}:
\begin{equation}\label{eq:max}
    a_{t:t+k-1}^{*} = \argmax_{\hat{a}_{0:k-1}^{m,n}}\hat{G}(\hat{s}_{0:k}^{m,n}, \hat{a}_{0:k-1}^{m,n})
\end{equation}

Compared to SOTA methods, our approach leverages dynamics model uncertainty by sampling $M$ individual VB weight candidates and maintaining a constant weight for each trajectory throughout its entire duration. In contrast, methods like COPlanner~\citep{wang2023coplanner}, P2P~\cite{wu2022plan}, and Plan2Explore~\cite{sekar2020planning} consider varying uncertainty values at each step, which propagates model uncertainty as the lookahead steps increase, potentially leading to misleading reward estimations.
Moreover, while our paper primarily focuses on learning uncertainty-aware dynamic models using Bayesian RL and acquiring exploratory policies with RND, our approach can also be applied using other uncertainty techniques, such as bootstrapped model ensemble methods in MBRL and uncertainty-oriented exploration methods in MFRL.

\subsubsection{Why Lookahead}\label{sec:why}
MBRL approximates a dynamics model based on the data collected in the environment. One way to leverage this model is by searching for an action sequence that maximizes the cumulative reward through entire trajectory optimization. However, a limitation of this approach is the accumulation of errors in the model, which hampers its effectiveness when the planning horizons are pretty long.

To tackle this challenge and enable efficient planning over long horizons, one approach is to refine the planning steps to $k$ and integrate a terminal state value function into the planning trajectory. Specifically, given a value function $V_\sigma$, we derive a policy $\pi_\phi^{k,V}$ by maximizing the k-step lookahead objective:
$$
\pi_\phi^{k,V}(s_0) = \argmax\limits_{a_0}\mathbb{E}_{\pi_\phi}[\mathbb{E}_{f_\theta}[\sum_{t=0}^{k-1}\gamma^{t} f_{\theta_r}(s_t,a_t)+\gamma^kV(s_{k}) ]],
$$ where $\pi_\phi^{k,V}$ indicates that the policy is based on k-step planning with value function $V$.

The performance of the overall policy depends on the quality of both the dynamics model $f_\theta$ and the value function $V_\sigma$. To illustrate the benefits of combining model-based k-step rollouts with the value function, we conduct an analysis and provide performance bounds for the k-step lookahead policy $\pi_\phi^{k,V}$ compared to its all-step horizon counterpart, which doesn't utilize the value function $V_\sigma$, as well as the one-step greedy policy derived from the value function alone.

\begin{theorem}
(k-step lookahead policy) Suppose $f_\theta$ is an uncertainty-aware dynamics model with uncertainty variation for all states bounded by $\epsilon_f$. Let $V_\sigma$ be an approximate value function for extrinsic rewards satisfying $\max\limits_s|V_\sigma^{*}(s)-V_\sigma(s)| \le \epsilon_v$, where $V_\sigma^{*}(s)$ is the optimal value funtion. Let the reward function $R(s, a)$ be bounded by $[0,R_{\max}]$, and $V_\sigma$ be bounded by $[0,V_{\max}]$. Let $\epsilon_p$ represent the distance from the suboptimality to the global optimum incurred in k-step lookahead optimization, such that $J^{\pi_\phi^*} - J^{\pi_\phi^{k,V}} \le \epsilon_p$, where $J^{\pi_\phi^*}$ is the global optimal return for the k-step optimization and $J^{\pi_\phi^{k,V}}$ is the results of the suboptimal k-step optimization. Then, the performance of the k-step lookahead policy $\pi_\phi^{k,V}$ can be bounded as follows:
\begin{equation}
\begin{aligned}
    &J^{\pi_\phi^*} - J^{\pi_\phi^{k,V}} \le \\
     &\frac{2}{1-\gamma^k}[R_{max}\sum_{t=0}^{k-1}\gamma^tt\epsilon_f+\gamma^kk\epsilon_fV_{max} +\frac{\epsilon_p}{2}+\gamma^k\epsilon_v].
\end{aligned}
\end{equation}

\end{theorem}

Intuited by Theorem 1, we can view the all-step horizon policy as a special case of $\pi_\phi^{k,V}$, where $k \to \infty$ and $V_\sigma(s) = 0$. The bound is primarily affected by the  dynamics model uncertainty variation $\epsilon_f$. This suggests that reducing the dynamics model uncertainty leads to a smaller optimality gap. 
In contract, the one-step greedy policy (k=1) is only affected by the value function approximation error $\epsilon_v$.
Moreover, comparing the k-step lookahead policy to the one-step greedy policy (k=1), we observe that the performance of the k-step lookahead policy is less affected by the value function approximation error $\epsilon_v$ and more influenced by the model uncertainty variation $\epsilon_f$. In fact, the dependency on $\epsilon_v$ decreases by at least a factor of $\gamma^{k-1}$. This implies that the k-step lookahead policy is beneficial when the value function approximation error exceeds the uncertainty variation in the learned model. Our empirical experiments in Sec.~\ref{sec:exp_lookahead} demonstrate that by carefully selecting an appropriate $k$, we can enhance policy performance in our framework.

\begin{figure*}[t]
    \centering
    \hfill
     \begin{subfigure}[t]{0.325\textwidth}
         \centering
         \includegraphics[width=\textwidth]{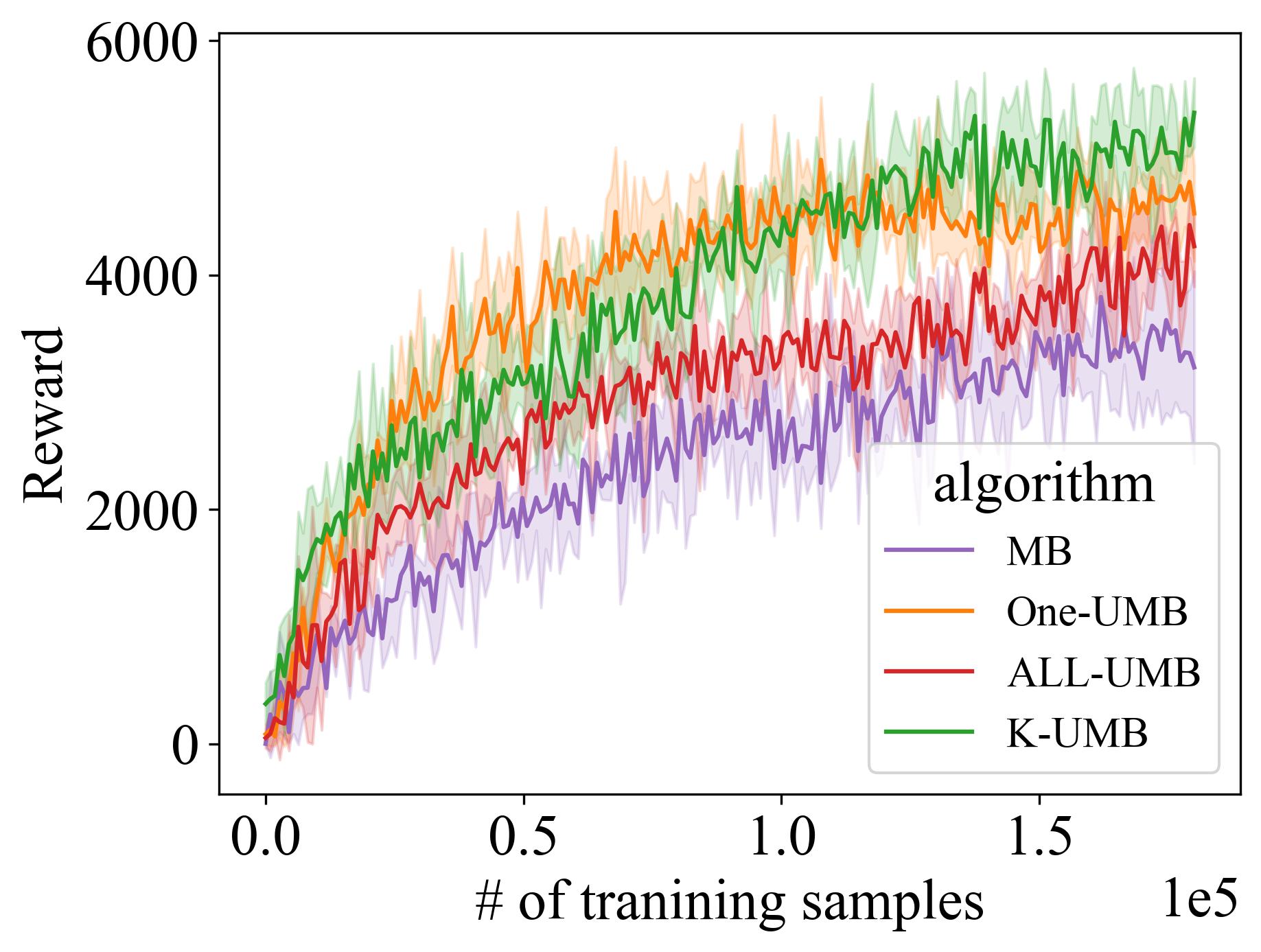}
         \caption{Walker2D}
         \label{fig:mb_walker2d}
     \end{subfigure}
     \hfill
     \begin{subfigure}[t]{0.325\textwidth}
         \centering
         \includegraphics[width=\textwidth]{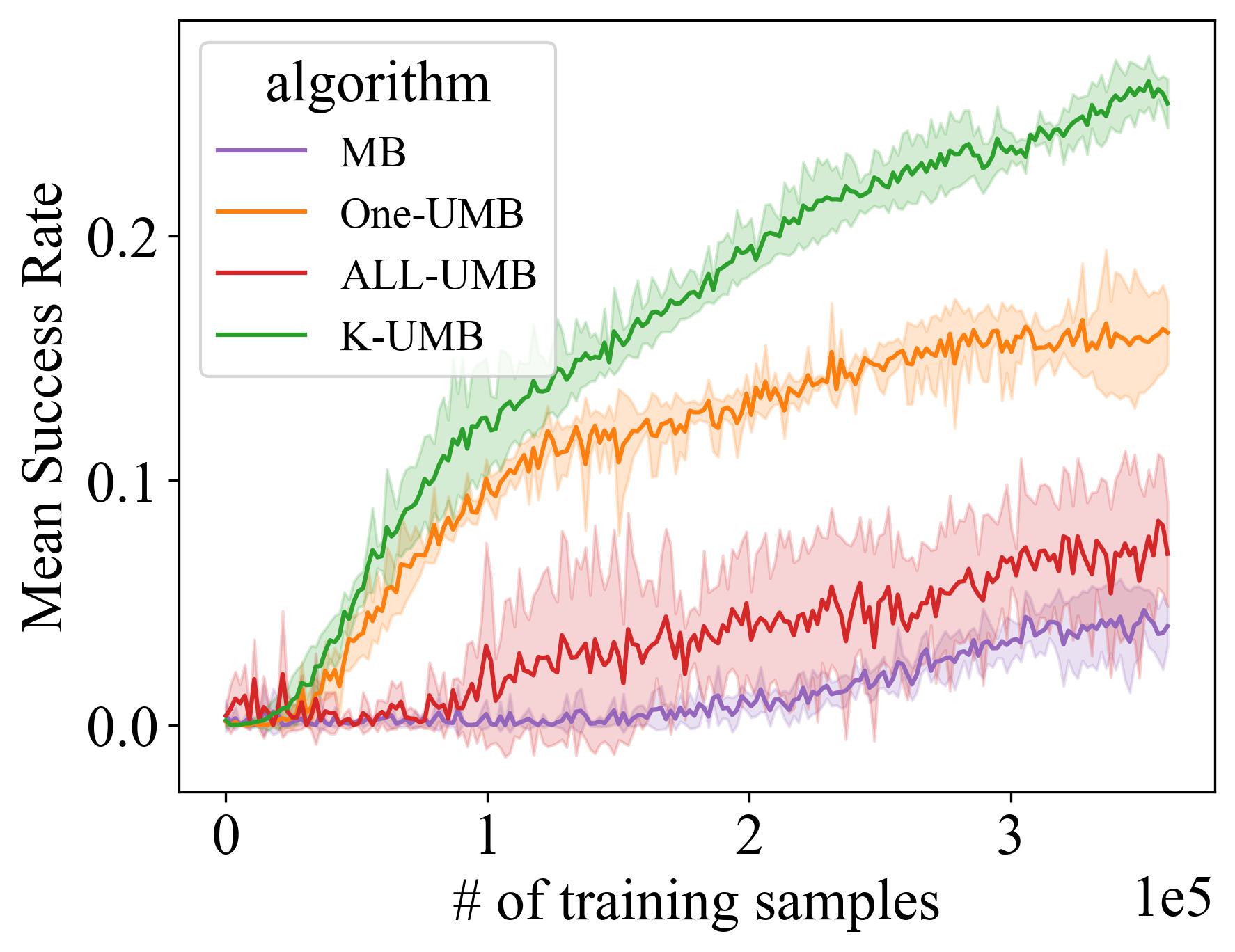}
         \caption{Manipulate}
         \label{fig:mb_hand}
     \end{subfigure}
    \hfill
    \begin{subfigure}[t]{0.325\textwidth}
         \centering
         \includegraphics[width=\textwidth]{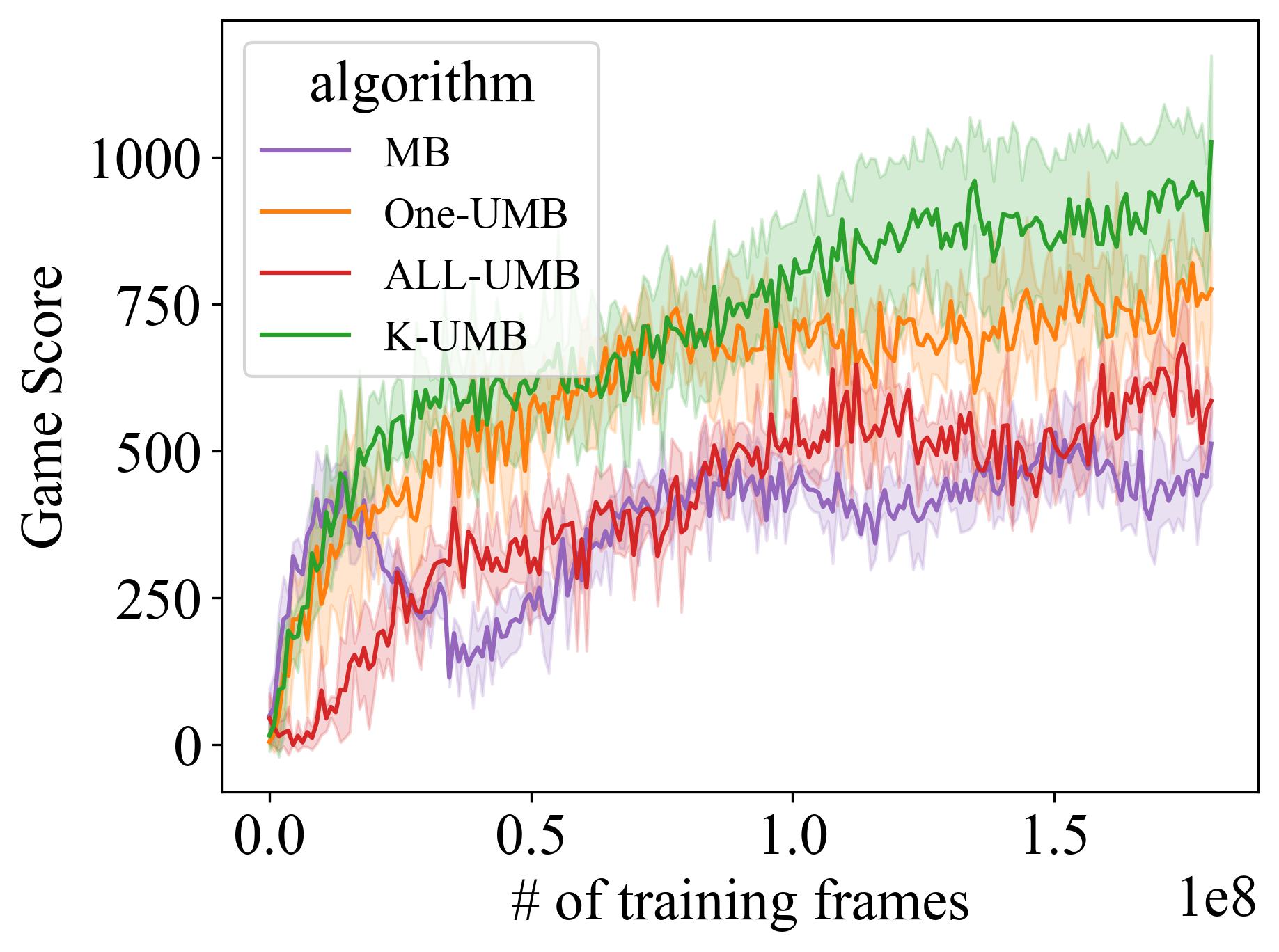}
         \caption{BeamRider}
         \label{fig:mb_BeamRider}
     \end{subfigure}
     \hfill
    \caption{The peformance of Walker2D, Hand Manipulate Block and BeamRider in different model-based settings.}
    \label{fig:mb}
 % \vspace{-0.2cm}
\end{figure*}

\section{Experiments}
\label{sec:exp}
In this section, we will explore the following questions:
\begin{itemize}
    \item Can incorporating model-based k-step lookahead planning with uncertainty yield significant benefits?
    \item Does uncertainty-aware policy optimization with model-based exploratory planning improve the quality of the dynamic model and enhance policy performance?
\end{itemize}

\subsection{Tasks}
Our method is highly versatile and can be applied to a wide range of RL tasks, regardless of the dimensionality of their states, the discreteness or continuity of their action spaces, and the sparsity or density of their rewards. To demonstrate its efficacy, we conducted empirical studies on control tasks using both the MuJoCo physics engine~\citep{todorov2012mujoco} and the Arcade Learning Environment (ALE)~\citep{bellemare2013arcade}.

The MuJoCo control tasks feature low-dimensional states but a continuous action space, while the ALE tasks involve high-dimensional image-based states and a discrete action space. We evaluated our method on the MuJoCo control tasks with both dense rewards (such as Walker2D, HalfCheetah, and Swimmer, etc.) and sparse rewards (such as Hand Manipulate Block, etc.). Additionally, we tested our approach on a suite of Atari games, including games with dense rewards (such as Beam Rider, Atlantis, and Freeway, etc.) and games with sparse rewards (such as Montezuma's Revenge, Gravitar, and Venture, etc.).  Fig.~\ref{fig:mb} and~\ref{fig:mb_mf} show results on partial environments. 

\subsection{Baselines}
To assess the effectiveness of our model-based k-step lookahead planning with uncertainty, we compared our method (K-UMB) against three baselines in Fig.~\ref{fig:mb}. The first baseline, MB, represents the standard MBRL approach without considering uncertainty~\citep{wang2019benchmarking,kaiser2019model}. The second baseline, one-UMB, utilizes uncertain MBRL with Bayesian RL technique~\citep{mcallister2016improving} and employs a one-step greedy policy.  Lastly, the third baseline, ALL-UMB, applies uncertain MBRL with an all-step horizon policy. 
% Lastly, the fourth baseline, ALL-UMB(op), involves uncertain MBRL with an all-step optimization policy that performs joint optimization for all horizon steps.
In practise, given the challenge of involve all-step horizon for tasks with a large horizon, we adopted a strategic approach. Instead of considering the complete all-step horizon, we employ a more efficient method by utilizing a $10*k$ step horizon.

\begin{figure*}[t]
    \centering
    \hfill
     \begin{subfigure}[t]{0.243\textwidth}
         \centering
         \includegraphics[width=\textwidth]{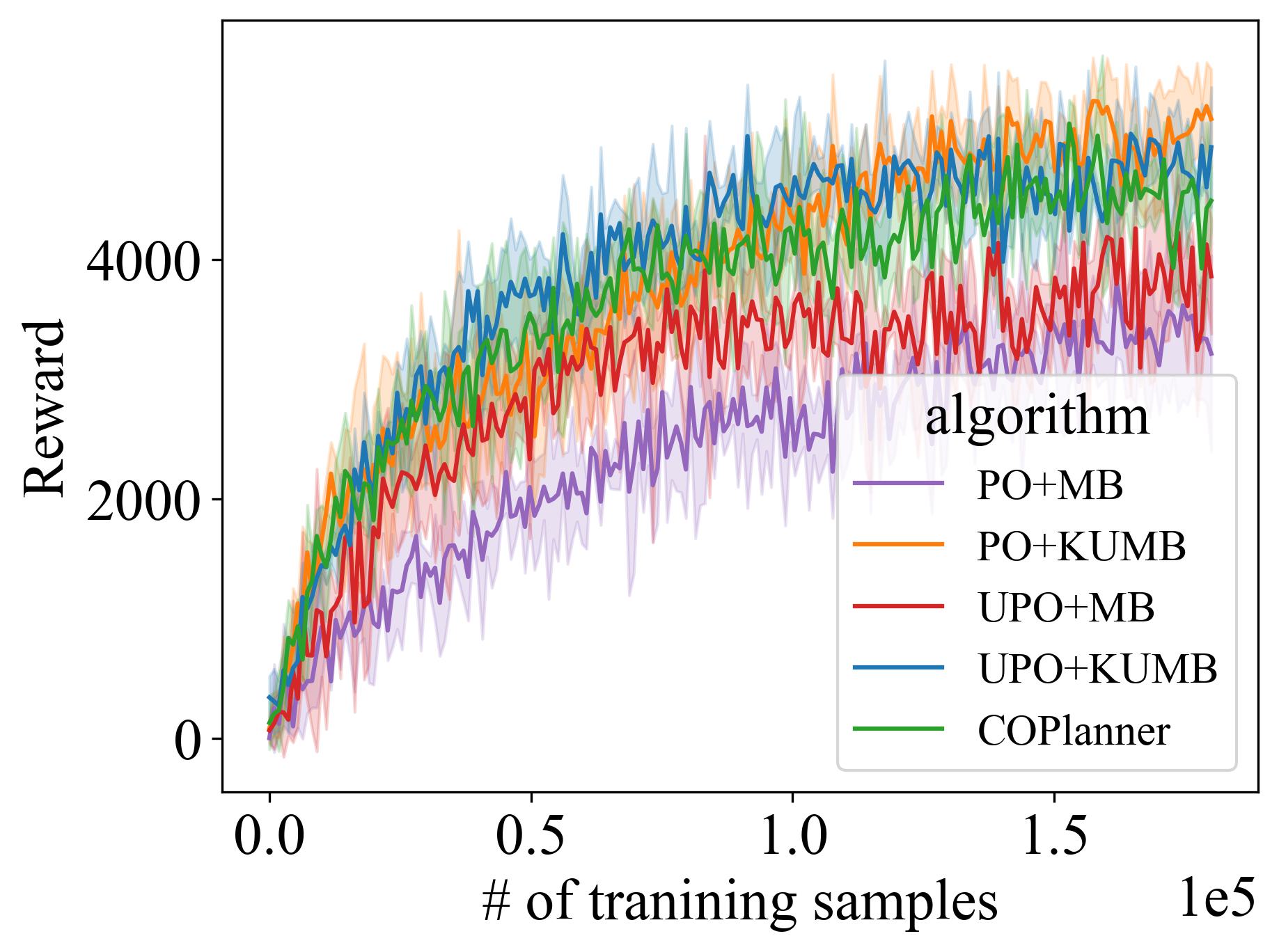}
         \caption{Walker2D}
         \label{fig:walker2d}
     \end{subfigure}
     \hfill
     \begin{subfigure}[t]{0.243\textwidth}
         \centering
         \includegraphics[width=\textwidth]{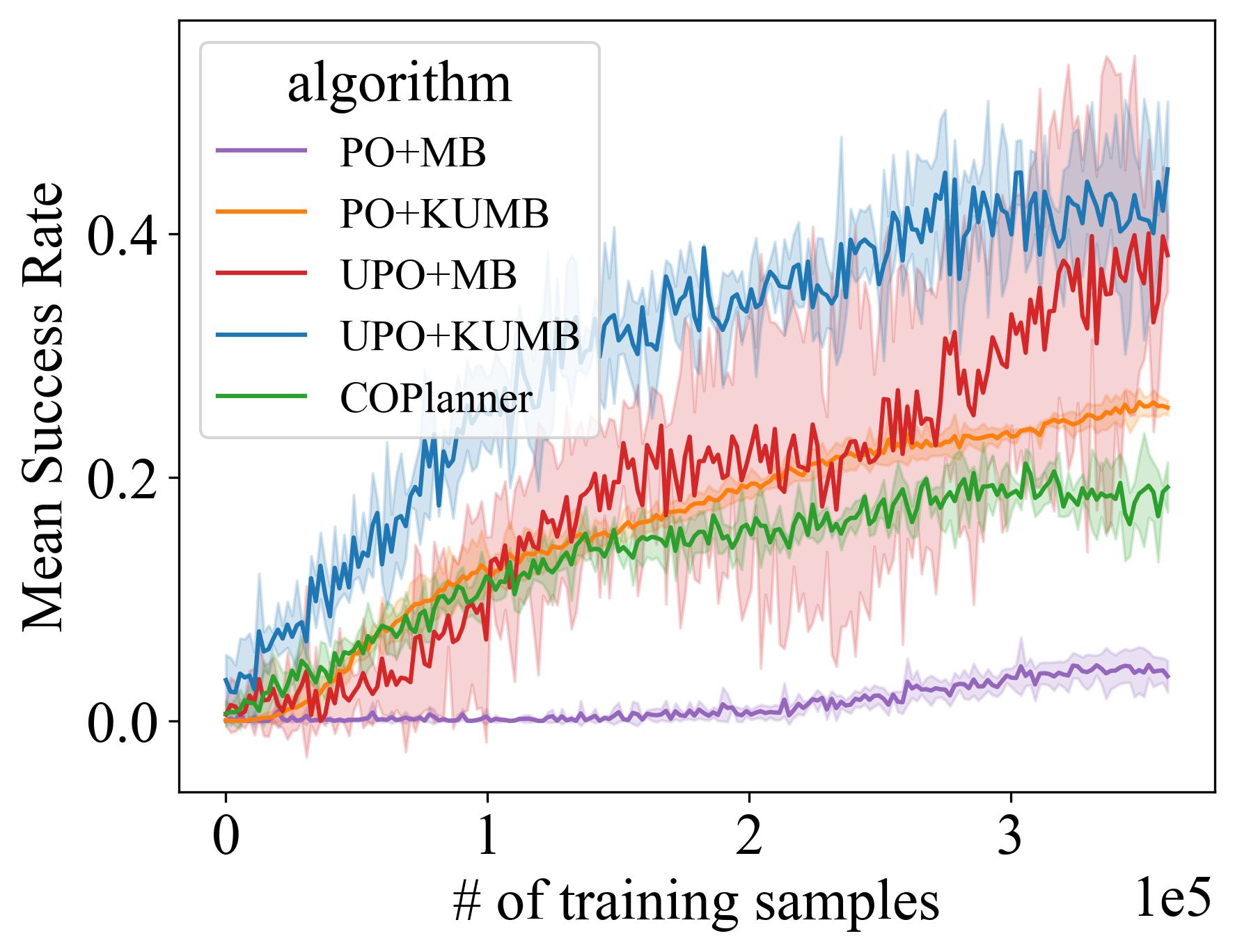}
         \caption{Manipulate Block}
         \label{fig:hand}
     \end{subfigure}
    \hfill
    \begin{subfigure}[t]{0.243\textwidth}
         \centering
         \includegraphics[width=\textwidth]{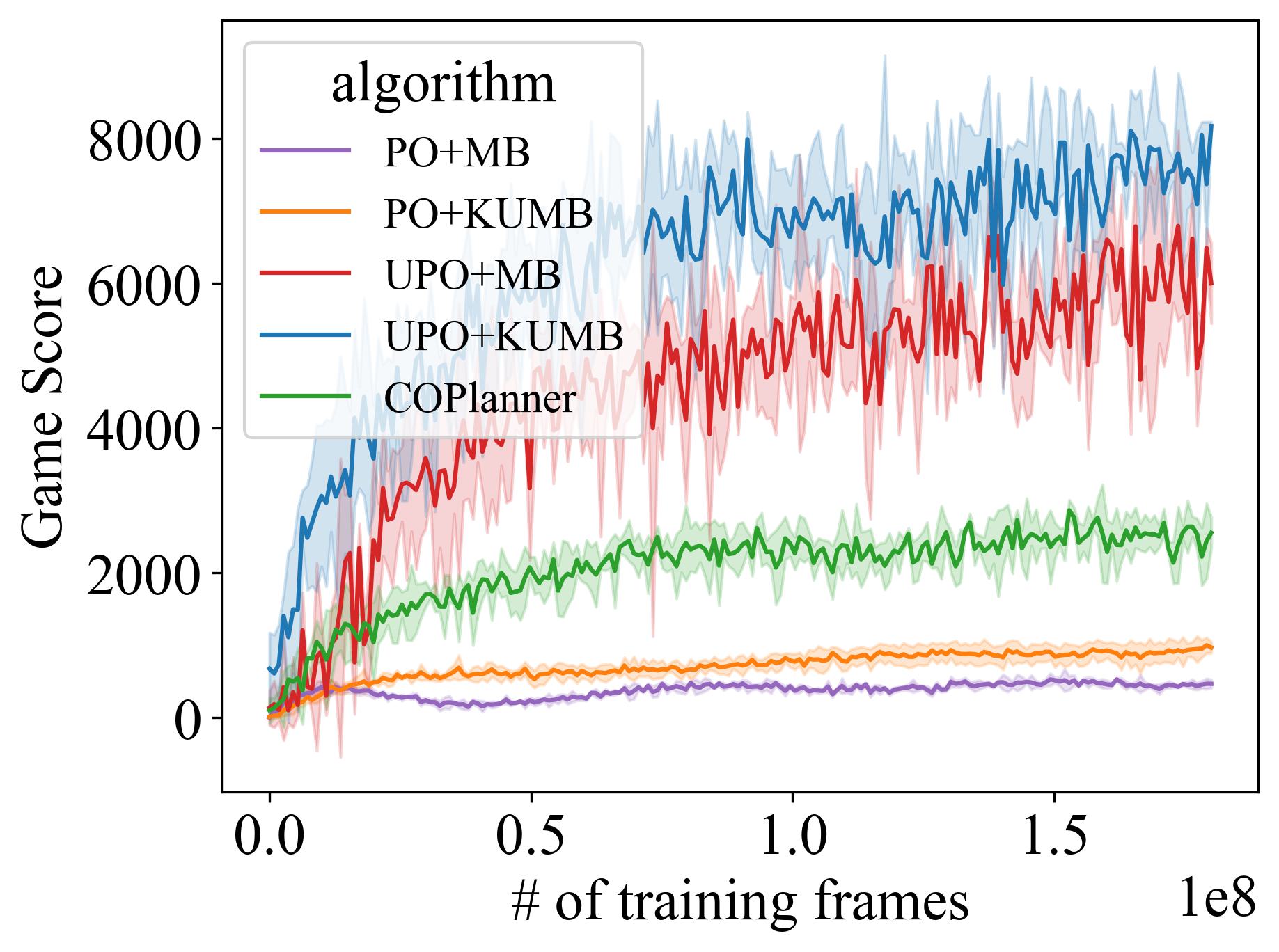}
         \caption{BeamRider}
         \label{fig:BeamRider}
     \end{subfigure}
     \hfill
     \begin{subfigure}[t]{0.243\textwidth}
         \centering
         \includegraphics[width=\textwidth]{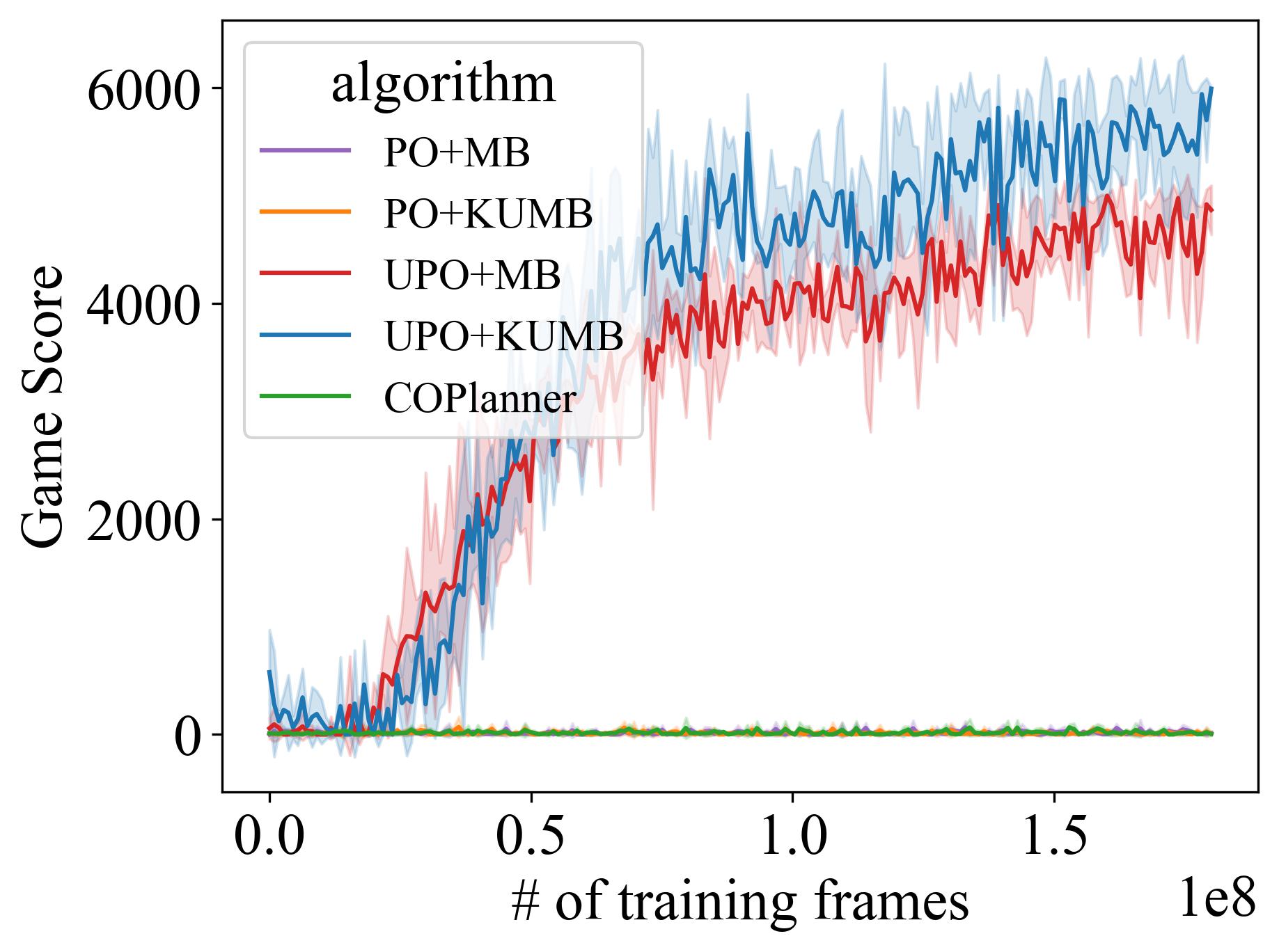}
         \caption{MR}
         \label{fig:mr}
     \end{subfigure}
     \hfill
    \caption{The performance of Walker2D, Hand Manipulate Block, BeamRider and Montezuma’s Revenge (MR) in different uncertainty exploration settings.}
    \label{fig:mb_mf}
    % \vspace{-0.2cm}
\end{figure*}

To further evaluate the benefits of incorporating uncertainty exploration in both the model-based planning (i.e., k-step lookahead uncertainty with VB dynamic model) and the policy optimization (i.e., RND~\citep{burda2018exploration}) phases, we compared our method (UPO+KUMB) against three baselines in Fig.~\ref{fig:mb_mf}: PO+MB (no uncertainty exploration in either phase), PO+KUMB (k-step uncertainty exploration only in the model-based planning phase), and UPO+MB (uncertainty exploration only in the policy optimization phase). We also compare our work with the state-of-the-art methods, such as COPlanner~\citep{wang2023coplanner} in Fig.~\ref{fig:mb_mf}, which learn dyna-style MBRL policy considering uncertainty.

 Moreover, we used the PPO~\citep{schulman2017proximal} algorithm as the base RL method for all the learning methods in our experiments. Each experiment was conducted 5 times with different random seeds, and we report the average performance with the standard deviation indicated by the shaded area.

 \begin{figure*}[t]
 \hfill
 \begin{minipage}[t]{0.325\textwidth}
    \centering
    \includegraphics[width=1\linewidth]{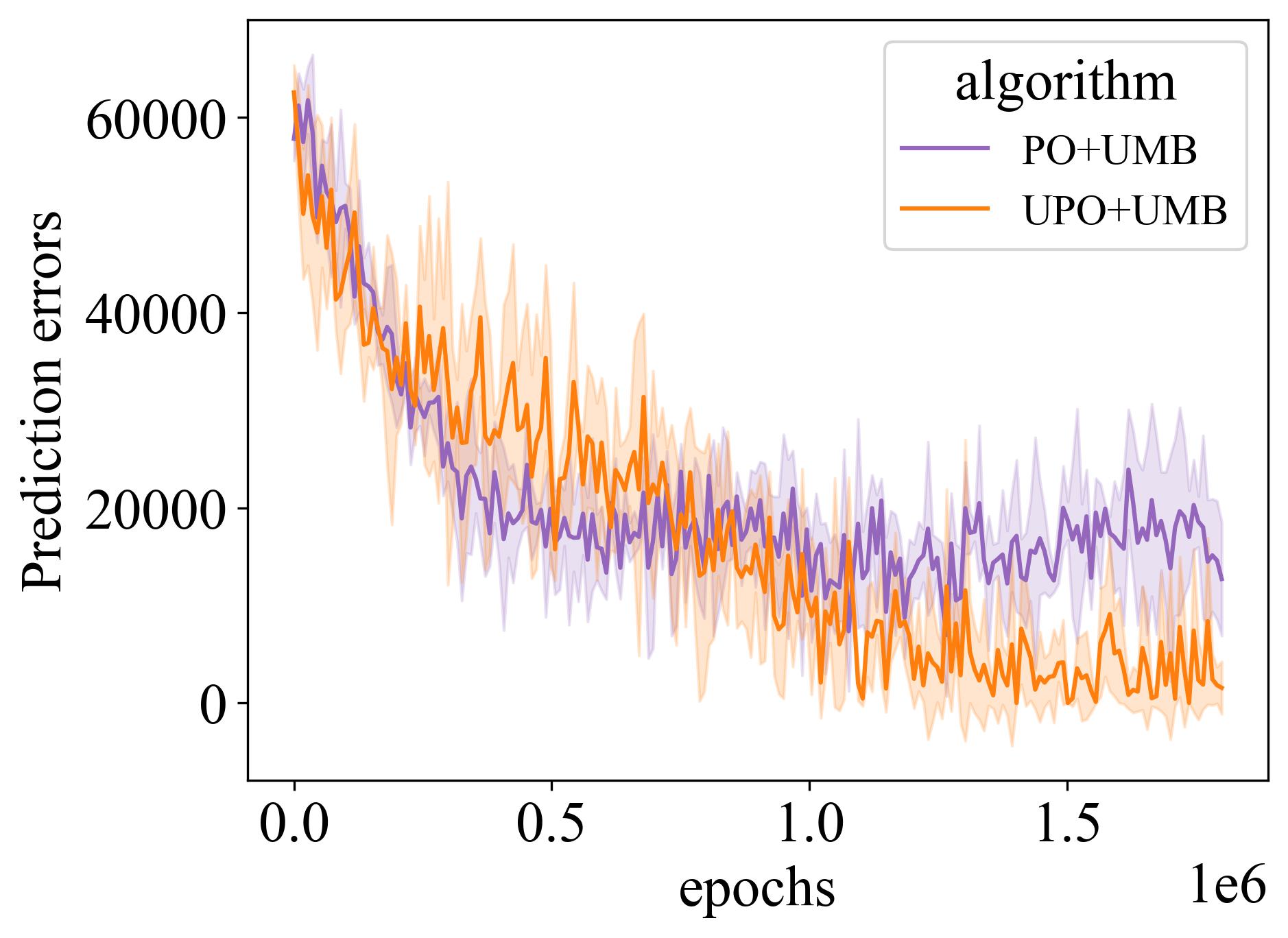}
    \caption{The dynamic model prediction errors in Montezuma’s Revenge (MR)}
    \label{fig:learning}
\end{minipage}
 \hfill
 \begin{minipage}[t]{0.325\textwidth}
    \centering
    \includegraphics[width=1\linewidth]{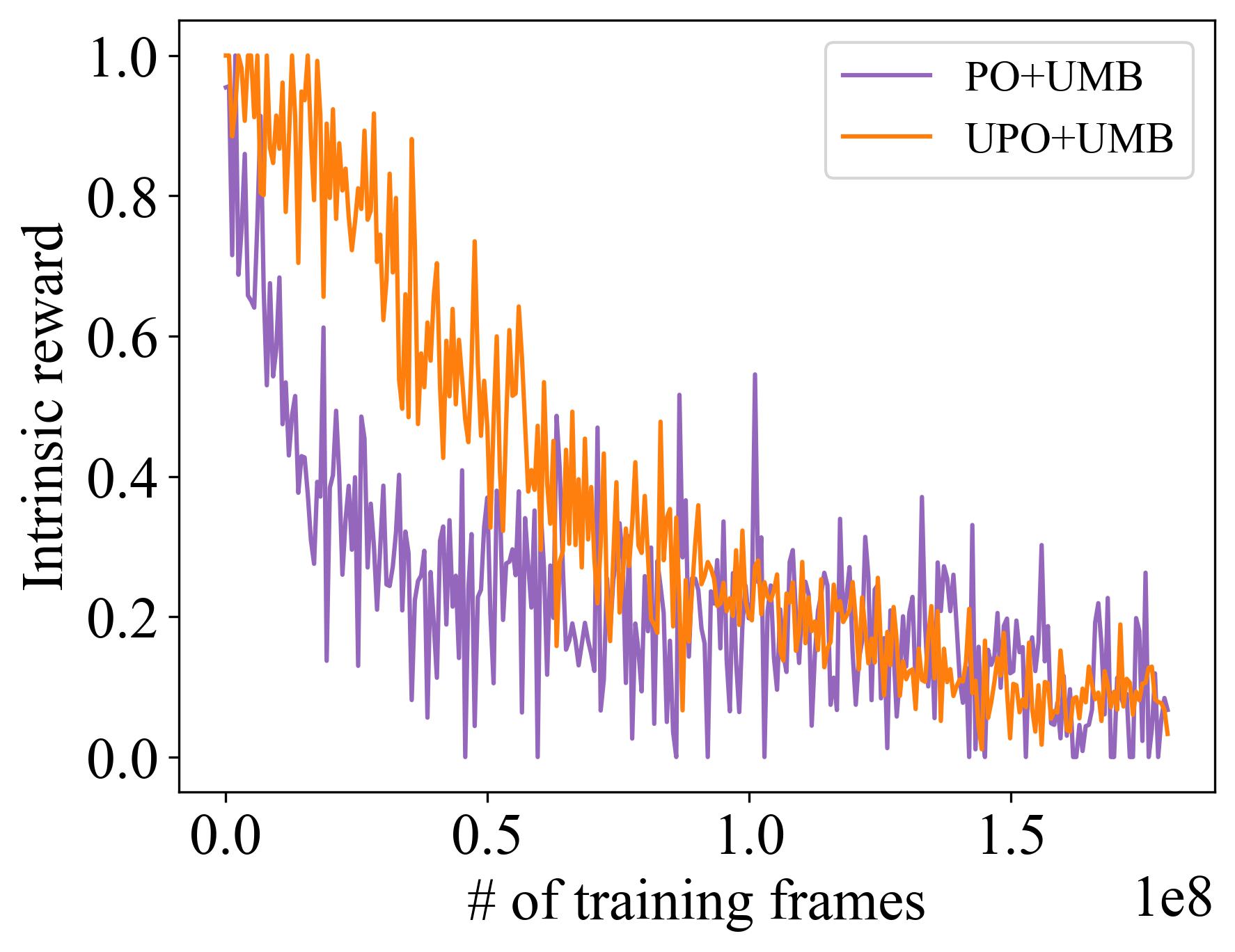}
    \caption{The uncertainty-exploration intrinsic reward (e.g., RND) in MR}
    \label{fig:ir}
\end{minipage}
 \hfill
    \begin{minipage}[t]{0.325\textwidth}
    \centering
    \includegraphics[width=1\linewidth]{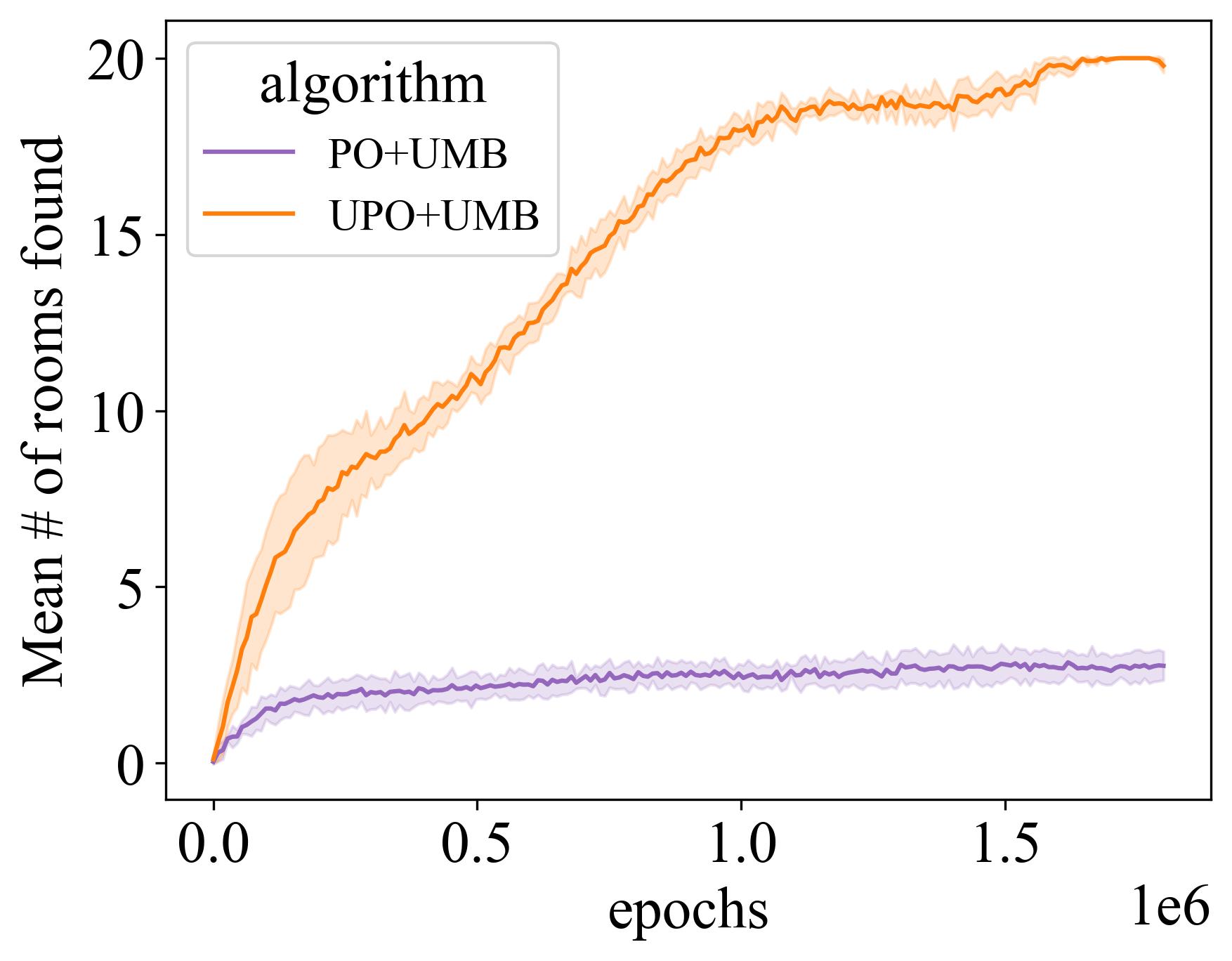}
    \caption{The mean number of rooms found in MR}
    \label{fig:room}
    \end{minipage}
 \hfill
 % \vspace{-0.3cm}
\end{figure*}

\subsection{Performance}
\subsubsection{K-step Lookahead with Uncertainty}\label{sec:exp_lookahead}
In Fig.~\ref{fig:mb}, we present the results of our k-step uncertainty lookahead method applied to three different environments: Walker2D, Hand Manipulate Block, and Beam Rider. These environments were chosen to represent continuous and discrete control tasks with dense and sparse rewards. Our k-UMB method consistently outperforms the three baseline methods across all environments. The standard MBRL method performs the worst as it does not consider the uncertainty in the dynamic model, leading to inaccurate results. The one-step greedy policy performs better than the standard MBRL method but worse than our k-UMB method. The ALL-UMB method perform worse than our k-UMB method due to the accumulation of uncertainty in the dynamic model approximation. Our results indicate that the k-step uncertainty lookahead approach effectively balances the one-step greedy policy and the all-step horizon/optimization policy, achieving optimal performance with an appropriately chosen value of k through grid search.  These findings emphasize the effectiveness of incorporating uncertainty in model-based planning and the importance of carefully choosing the horizon for optimal results.

For tasks like Montezuma’s Revenge, which involve complex image states and sparse rewards, achieving a non-zero score based solely on uncertainty-aware model-based planning is challenging, as shown in Fig.~\ref{fig:mr}. Therefore, incorporating uncertainty exploration in policy optimization phase is necessary, as discussed below.

\subsubsection{Uncertainty exploration in planning and policy optimization}
In this experiment, we examine the benefits of incorporating uncertainty-driven exploration in both model-based planning and policy optimization. Our goal is to determine if using uncertainty-based exploration during policy optimization (e.g., with RND) to collect training data can improve the quality of the learned dynamics model, particularly in tasks with sparse rewards.
Figures~\ref{fig:learning},~\ref{fig:ir} and~\ref{fig:room} compare model prediction error, intrinsic reward, and the number of rooms found between standard policy optimization (PO) and uncertainty-explored policy optimization (UPO) in Montezuma’s Revenge during the training process. Notably, in Fig.\ref{fig:learning}, the model prediction error is evaluated using the mean squared error of image state predictions. In Fig.\ref{fig:ir}, an RND model is also trained for the PO agent, serving exclusively to compute the intrinsic reward without influencing the learning process.
From Fig.~\ref{fig:learning}, it is evident that UPO eventually leads to lower prediction errors compared to PO. In the early stage, the UPO agent tend to explore more rooms with diverse new states, causing its prediction error and intrinsic reward (Fig.~\ref{fig:ir}) to converge more slowly and with higher value than PO. The PO agent converges faster but tends to explore a smaller, localized number of rooms (Fig.~\ref{fig:room}), quickly becoming familiar with those states.
As training progresses, UPO explores and becomes familiar with more rooms. By the time of convergence, the PO agent remains confined to local rooms. Though the PO agent occasionally encounters unfamiliar states, the number of these new states is insufficient to boost game score due to their sparse rewards nature, resulting in relative large mean prediction errors and intrinsic reward variance. Conversely, the UPO agent, familiar with a larger number of rooms, has fewer opportunities to encounter new states, leading to smaller prediction errors and intrinsic reward variance.
These findings indicate that uncertainty-driven exploration in policy optimization ultimately provides more informative and diverse data, enhancing the dynamics model's performance.

Furthermore, we examine the impact of incorporating uncertainty in both sides on task performance. In Fig.~\ref{fig:mb_mf}, we demonstrate that uncertainty exploration in policy optimization significantly improves performance in tasks with sparse rewards and high-dimensional complex states, such as Montezuma's Revenge, Hand Manipulate Block and Beam Rider, and slightly improves performance in tasks with dense rewards and low-dimensional states, such as Walker2D. The uncertainty exploration in policy optimization helps the agent gather more informative samples with positive rewards, particularly in tasks with sparse rewards and complex observations. Additionally, our results show that the k-step uncertainty lookahead in model-based planning improves performance in all tasks, regardless of whether the policy optimization incorporates uncertainty exploration or not. In summary, our findings suggest that incorporating uncertainty in both model-based planning and policy optimization can enhance learning in a variety of environments, especially those with sparse rewards, leading to more robust and effective performance.
In comparison to COPlanner, which surpasses the recent work~\citep{wu2022plan}, our approach consistently demonstrates superior performance. While the results on Walder2D exhibit similarities, our method exhibits a significant advantage in the remaining three scenarios. We hypothesize that COPlanner's suboptimal performance in sparse reward and high-dimensional state settings may be attributed to challenges in model uncertainty propagation, potentially resulting in misleading reward estimations.

% This highlights the importance of considering uncertainty in the learning process and opens up opportunities for further exploration of how to integrate uncertainty in different RL approaches.

% \section{Discussion}\label{sec12}

% Discussions should be brief and focused. In some disciplines use of Discussion or `Conclusion' is interchangeable. It is not mandatory to use both. Some journals prefer a section `Results and Discussion' followed by a section `Conclusion'. Please refer to Journal-level guidance for any specific requirements. 

\section{Conclusion}\label{sec13}

In this study, we propose a novel framework for uncertainty-aware
policy optimization with model-based exploratory planning to improve both model accuracy and policy performance. Our framework incorporates k-step lookahead planning with uncertainty, revealing an inherent trade-off between model uncertainty and value function approximation error, while enhancing policy performance accordingly. Additionally, the utilization of an uncertainty-driven exploratory policy in the policy optimization phase facilitates the collection of diverse and informative training samples.
The empirical results demonstrate the effectiveness of our framework across a wide range of tasks, surpassing state-of-the-art methods while requiring fewer interactions. 

% \backmatter

% \bmhead{Supplementary information}

% If your article has accompanying supplementary file/s please state so here. 

% Authors reporting data from electrophoretic gels and blots should supply the full unprocessed scans for key as part of their Supplementary information. This may be requested by the editorial team/s if it is missing.

% Please refer to Journal-level guidance for any specific requirements.

% \bmhead{Acknowledgments}

% Acknowledgments are not compulsory. Where included they should be brief. Grant or contribution numbers may be acknowledged.

% Please refer to Journal-level guidance for any specific requirements.

% \noindent
% If any of the sections are not relevant to your manuscript, please include the heading and write `Not applicable' for that section. 

% %%===================================================%%
% %% For presentation purpose, we have included        %%
% %% \bigskip command. please ignore this.             %%
% %%===================================================%%
% \bigskip

% Springer journals and proceedings: \url{https://www.springer.com/gp/editorial-policies}

% Nature Portfolio journals: \url{https://www.nature.com/nature-research/editorial-policies}

% \textit{Scientific Reports}: \url{https://www.nature.com/srep/journal-policies/editorial-policies}

% BMC journals: \url{https://www.biomedcentral.com/getpublished/editorial-policies}
% \end{flushleft}

% \appendix

\bibliographystyle{named}
\bibliography{ijcai22}

\end{document}